\documentclass[letterpaper, 10 pt, conference]{ieeeconf}  

\IEEEoverridecommandlockouts                              

\usepackage{graphics} 
\usepackage{epsfig} 
\usepackage{mathptmx} 
\usepackage{times} 
\usepackage{amsmath} 
\usepackage{amssymb}  
\usepackage{float}
\usepackage{makecell}
\usepackage{multirow}
\usepackage{caption}
\usepackage{titlesec}
\usepackage{url}
\usepackage{stfloats}
\usepackage{cite}
\usepackage{booktabs}
\usepackage{hhline}
\usepackage{array}
\usepackage[table]{xcolor}

\titleformat{\subsubsection}[runin]
{\bfseries}{}{0pt}{}[.\hspace{0.5em}   ]
\titlespacing{\subsubsection}{0pt}{0.5em}{0pt}

\title{\LARGE \bf
Walking with Terrain Reconstruction: \\Learning to Traverse Risky Sparse Footholds
}

\author{Ruiqi Yu$^{1}$ Qianshi Wang$^{1}$ Yizhen Wang$^{1}$ Zhicheng Wang$^{1}$ Jun Wu$^{1,2}$ Qiuguo Zhu$^{*1,2}$
\thanks{This work was supported by the National Key R\&D Program of China (Grant No. 2022YFB4701502), the ”Leading Goose” R\&D Program of Zhejiang (Grant No. 2023C01177), and the 2035 Key Technological Innovation Program of Ningbo City (Grant No. 2024Z300).}
\thanks{$^{1}$The authors are with Institute of Cyber-Systems and Control, Zhejiang University, 310027, China.}%
\thanks{$^{2}$Qiuguo Zhu and Jun Wu are with State Key Laboratory of Industrial Control Technology, 310027, China.}%
\thanks{$^*$ Qiuguo Zhu ({\tt\small qgzhu@zju.edu.cn}) is the corresponding author.}%
}

\begin{document}

\maketitle
\thispagestyle{empty}
\pagestyle{empty}

\begin{abstract}

Traversing risky terrains with sparse footholds presents significant challenges for legged robots, requiring precise foot placement in safe areas. To acquire comprehensive exteroceptive information, prior studies have employed motion capture systems or mapping techniques to generate heightmap for locomotion policy. However, these approaches require specialized pipelines and often introduce additional noise. While depth images from egocentric vision systems are cost-effective, their limited field of view and sparse information hinder the integration of terrain structure details into implicit features, which are essential for generating precise actions. In this paper, we demonstrate that end-to-end reinforcement learning relying solely on proprioception and depth images is capable of traversing risky terrains with high sparsity and randomness. Our method introduces local terrain reconstruction, leveraging the benefits of clear features and sufficient information from the heightmap, which serves as an intermediary for visual feature extraction and motion generation. This allows the policy to effectively represent and memorize critical terrain information. We deploy the proposed framework on a low-cost quadrupedal robot, achieving agile and adaptive locomotion across various challenging terrains and showcasing outstanding performance in real-world scenarios.
Video at: \url{youtu.be/Rj9v5EZsn-M}.

\end{abstract}

\section{INTRODUCTION}

Quadrupedal robots have attracted significant attention for their ability to navigate diverse terrains by adjusting their foot placements freely. To fully leverage their mobility advantages in complex environments, these robots face the challenge of operating on risky and complex terrains with sparse footholds, such as stepping stones, balance beams, and gaps. To traverse such terrains stably and flexibly, robots must acquire precise environmental awareness and adapt their gait dynamically, while maintaining balance and stability. They must plan and execute each step with precision to ensure that each foot lands safely within the safe area.

\begin{figure}[!htbp]
  \centering
  \includegraphics[width=1.0\linewidth]{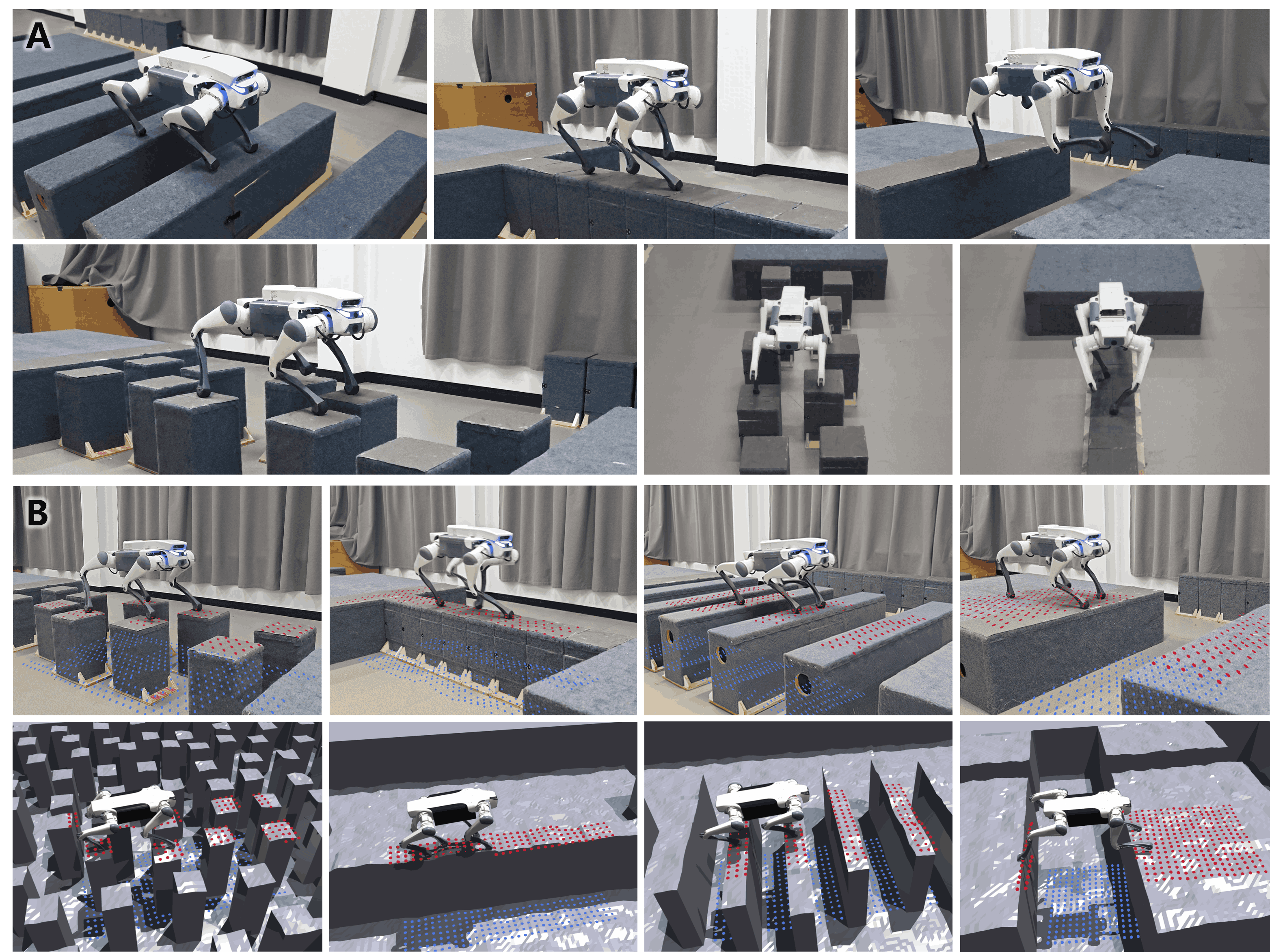}
  \vspace{-13pt}
  \caption{The proposed framework employs local terrain reconstruction to extract comprehensive and detailed terrain information from limited visual perception, significantly enhancing the policy's environmental understanding. (A) Using end-to-end RL, it enables agile locomotion across risky terrains with sparse footholds, such as stepping stones, balance beams, stepping beams and gaps, showcasing remarkable flexibility and adaptability in the real world. (B) Reconstructed local heightmap in the real world and simulation.}
  \label{fig1}
  \vspace{-18pt}
\end{figure}

Legged locomotion on risky terrains with sparse footholds relies on the seamless integration of perception and control. Existing studies often utilize model-based hierarchical controllers \cite{griffin2019footstep,mastalli2020motion,fahmi2022vital}, decomposing the problem into multiple stages of perception, planning, and control \cite{yu2022visual}. Robots typically use SLAM for heightmap generation and employ nonlinear optimization \cite{winkler2018gait} or graph search \cite{griffin2019footstep} to solve motion trajectories, which are then tracked by low-level controllers \cite{margolis2022learning,yu2022visual,villarreal2020mpc,gangapurwala2022rloc}. These methods require extensive manual design to adapt agile locomotion to specific scenarios \cite{margolis2022learning,yu2022visual}, and the trade-off between model accuracy and computational cost limits their scalability for more complex terrains \cite{margolis2022learning,gangapurwala2022rloc}. Some studies have also integrated Reinforcement Learning (RL) in planning or control, such as generating base trajectories \cite{xie2022glide} or foot placements \cite{yu2022visual,gangapurwala2022rloc}. \cite{jenelten2024dtc} applied Trajectory Optimization (TO) for foot placement generation, followed by RL-based tracking, demonstrating impressive results in complex environments. However, such decoupled architectures may restrict the performance of each module, leading to conservative actions \cite{margolis2022learning,zhang2023learning}.

Recently, end-to-end RL approaches have shown strong capabilities for traversing various complex terrains \cite{miki2022learning,agarwal2023legged,cheng2024extreme,yang2021learning,yang2023neural,zhuang2023robot,zhang2023learning,hoeller2024anymal,luo2024pie}, yielding remarkable results in unstructured environments, and have been explored to tackle sparse footholds. The main challenge for end-to-end RL in such scenarios lies in obtaining accurate and comprehensive perception of the surrounding environment. Heightmap offers a memory-efficient and simple yet physically explicit representation of terrain \cite{miki2022elevation}. It contains necessary geometric information for motion planning in risky terrains and is widely used as exteroceptive perception for locomotion \cite{miki2022learning,zhang2023learning,hoeller2024anymal}, though it still have certain limitations. For example, \cite{zhang2023learning} demonstrated agile locomotion with an end-to-end RL policy using heightmap in highly risky terrains. However, it depends on motion capture systems for state estimation during real-world deployment, which limits its applicability. Model-based heightmap reconstruction methods \cite{fankhauser2018probabilistic} build heightmap by combining data from odometry and cameras or LiDAR, but are susceptible to localization drift \cite{hoeller2022neural}. Learning-based approaches \cite{hoeller2022neural,hoeller2024anymal} have shown promising results in reconstructing local terrain from noisy multiview observations, but still rely on the robot's global pose to align consecutive frames. Additionally, these methods often requires a complicated hardware pipeline \cite{yang2023neural} and introduce additional computational overhead, making them unsuitable for low-cost robots.

In contrast, recent works have bypassed heightmap, utilizing vision encoding modules to extract environmental information from single-view depth images and infer the surrounding terrain implicitly, achieving great success
\cite{agarwal2023legged,cheng2024extreme,yang2021learning,yang2023neural,zhuang2023robot,luo2024pie}. However, this brings in new challenges. First, the limitations of viewpoint and potential occlusions create partially observable problems \cite{yang2023neural}, necessitating short-term memory of past observations to provide comprehensive local terrain information. Second, we find that the features implicitly extracted are insufficient to fully represent the necessary terrain details. This results in a tendency to learn a single gait pattern that averages across all sparsity and randomness. Can we develop a simple framework that provides sufficient and clearly-represented terrain features using only limited depth images, thus broadening the applicability of quadrupedal robots in risky sparse footholds?

In this paper, we propose a single-stage end-to-end RL-based framework to address these challenges. The architecture is divided into two key components: the terrain reconstructor and the locomotion policy. The memory-based terrain reconstructor utilizes proprioception and egocentric vision to reconstruct local terrain heightmap, including currently invisible areas beneath and behind the robot. The locomotion policy takes proprioception and the reconstructed heightmap as input to infer the robot's state and the surroundings with implicit-explicit estimation, enabling the policy to fully understand environmental features. Local terrain reconstruction introduces physically significant supervision targets for encoding information-sparse depth images, prompting the policy to extract comprehensive terrain features essential for safe navigation (e.g. stone positions and terrain edges), thereby facilitating exploration and reducing the difficulty of acquiring sparse reward in complex terrains. Moreover, it avoids global mapping and localization for heightmap generation, eliminating potential noise and drift \cite{fankhauser2018probabilistic}. We deploy the proposed framework on a low-cost quadrupedal robot and demonstrate agile locomotion on challenging terrains, such as stepping stones and balance beams, as shown in Fig. \ref{fig1}.

In summary, our main contributions include:

\begin{itemize}
\item \textbf{A single-stage end-to-end learning framework} that introduces \textbf{local heightmap as an intermediate representation} is proposed to significantly enhance environmental understanding and enable precise foot placements \textbf{using only proprioception and view-limited egocentric vision}.
\item \textbf{A terrain reconstructor module} is proposed to accurately reconstruct the local terrain, providing explicit supervision targets that encourage the policy to comprehensively extract and memorize exteroceptive information thus representing essential terrain features.
\item \textbf{Successful sim-to-real transfer} of agile locomotion on \textbf{risky terrains with sparse footholds} such as stepping stones, balance beams, stepping beams and gaps, demostrating exceptional adaptability and robustness.
\end{itemize}

\section{METHOD}

The proposed framework is based on end-to-end RL and employs a set of neural network models to generate target joint angle from raw depth images and proprioception, with local heightmap serving as an intermediate representation. The framework consists of a locomotion policy and a terrain reconstructor, as shown in Figure 2. All model components are jointly optimized in a single-stage training pipeline, requiring no need for additional training phase for the terrain reconstructor, and could subsequently be deployed to a real robot via zero-shot transfer.


\subsection{Terrain-Aware Locomotion Policy} 

The locomotion policy is composed of an I-E Estimator and an actor policy. Inspired by \cite{wang2024toward,luo2024pie}, the I-E Estimator is designed to estimate the robot's state and privileged environmental information implicitly or explicitly, while also reconstructing future proprioception. This enhances the policy's environmental awareness and adaptability. To mitigate potential data inefficiency and information loss introduced by the two-stage training paradigm \cite{fu2023deep,nahrendra2023dreamwaq,ji2022concurrent,luo2024pie}, we adopt an asymmetric actor-critic structure \cite{nahrendra2023dreamwaq}, simplifying the training into a single stage. The policy components are trained by optimizing the actor-critic architecture along with the I-E Estimator, with the actor-critic being optimized using the Proximal Policy Optimization (PPO) algorithm \cite{schulman2017proximal}.


\subsubsection{Observations and Action}

The actor policy takes as input the latest proprioception $o_t$, the estimated base velocity $\hat v_t \in \mathbb{R}^{3}$, encodings of heightmap around the four feet and body $\hat z^f_t \in \mathbb{R}^{16}$ and $\hat{z}^m_t \in \mathbb{R}^{16}$, and the latent vector $z_t$, to generate joint position actions $a_t \in \mathbb{R}^{12}$ for PD controllers to track. The proprioception $o_t \in \mathbb{R}^{45}$ is measured from the joint encoders and IMU and is defined as:
$$
\setlength{\abovedisplayskip}{3pt}
\setlength{\belowdisplayskip}{3pt}
o_t=[\omega_t\quad g_t\quad cmd_t\quad \theta_t\quad \dot\theta_t\quad a_{t-1}]^T \eqno{(1)}
$$
where $\omega_t$, $g_t$, $cmd_t$, $\theta_t$, $\dot\theta_t$, and $a_{t-1}$ are the body angular velocity, gravity vector in the body frame, velocity command, joint angle, joint angular velocity, and previous action, respectively. The critic policy receives not only $o_t$ but also privileged environmental information $s_t$, which is defined as:
$$
\setlength{\abovedisplayskip}{3pt}
\setlength{\belowdisplayskip}{3pt}
s_t=[v_t\quad c_t\quad H^b_t\quad H^f_t]^T \eqno{(2)}
$$
where $v_t$, $c_t$, $H^b_t$, $H^f_t$ are base velocity, foot-ground contact state, and the ground truth heightmap around the robot’s body and four feet, respectively.

\begin{figure*}[!htbp]
  \centering
  \vspace{3.5pt}
  \includegraphics[width=0.9\linewidth]{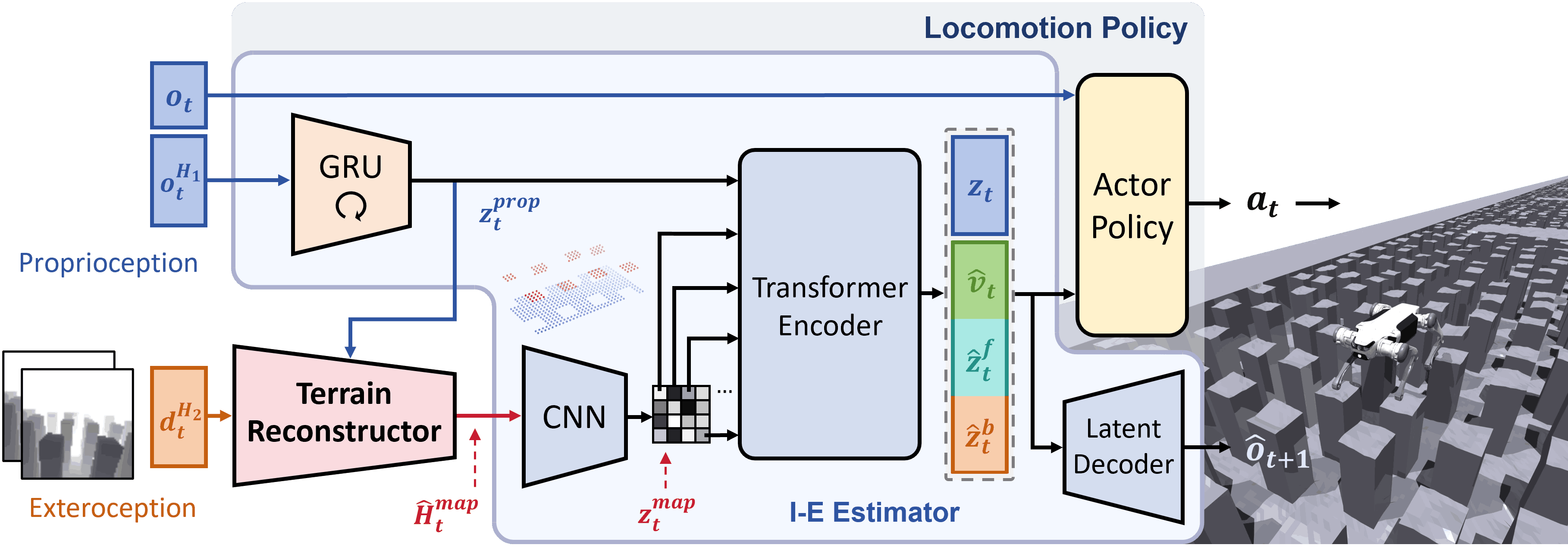}
  \caption{Overview of the proposed framework. The end-to-end framework consists of two key modules: a {locomotion policy} (gray) and a {terrain reconstructor} (pink). The terrain reconstructor takes depth images as input to generate local terrain heightmap reconstruction $\hat H^{\text{map}}_t$ while the locomotion policy performs implicit-explicit estimation and subsequently generates actions. All components are jointly optimized by PPO and supervised learning within the same training stage.}
  \vspace{-3pt}
  \label{fig2}
  \vspace{-13pt}
\end{figure*}

\subsubsection{Rewards}

The reward function basically follows the configurations in \cite{nahrendra2023dreamwaq,rudin2022learning,lee2020learning}. Additionally, to prevent the robot from easily stepping onto the edges of sparse terrains that may lead to falls in the real world, we introduce an additional reward term to penalize foot placements near terrain edges. This encourages the robot to consciously select actions that ensure each footstep lands in the safe area:
$$
\setlength{\abovedisplayskip}{5pt}
\setlength{\belowdisplayskip}{5pt}
r_{\text{feetedge}}=-\sum_{i\in \text{feet}}c_i\cdot\sum_{d\in \text{dist}} \omega_d \cdot E_d[p_i] \eqno{(3)}
$$
where $c_i$ is the contact state. $E_d$ is a boolean function indicating whether foot position $p_i$ is within the specified distance threshold $d$ from the edge. $\omega_d$ is the penalty coefficient for different distances, with larger penalties applied as the distance decreases. Complete rewards are listed in Table \ref{table1}, and check Appendix\ref{appendixA} for computational details of $r_{\text{feetedge}}$.

\begin{table}[!htbp]
\setlength{\abovecaptionskip}{0pt}
\setlength{\belowcaptionskip}{0pt}
\vspace{-3pt}
\caption{Detail reward terms.}
\vspace{-3pt}
\label{table1}
\begin{center}
\renewcommand\arraystretch{1.2}
\begin{tabular}{lll}
\Xhline{0.75pt}
{Reward Term}             & Equation                                 & Weight              \\
\hline
Lin. vel. tracking  & $\text{exp}(-||\text{min}(v, v^{\text{cmd}})-v^{\text{cmd}}||^2/0.25)$         & 1.5                 \\
Ang. vel. tracking & $\text{exp}(-||\omega_{yaw}-\omega^{\text{cmd}}_{\text{yaw}}||^2/0.25)$ & 0.5       \\
Lin. vel. (z)        & $v_z^2$                                  & -2.0                \\
Ang. vel. (xy)     & $||w_{xy}||^2$                           & -0.05               \\
Orientation               & $g_x^2+g_y^2$                            & -1.0                \\
Torques                   & $\sum_{j\in \text{joints}}|\tau_j|^2$           & -1.0$\times10^{-5}$ \\
Action rate               & $\sum_{j\in \text{joints}}|a_t-a_{t-1}|^2$                & -0.01               \\
Smoothness                & $\sum_{j\in \text{joints}}|a_t-2a_{t-1} + a_{t-2}|^2$     & -0.01               \\
Joint power               & $\sum_{j\in \text{joints}}|\tau_j||\dot{q_j}|$  & -2.0$\times10^{-5}$ \\
Joint accelerations       & $\sum_{j\in \text{joints}}|\ddot{q_j}|^2$       & -2.5$\times10^{-7}$ \\
Joint error               & $\sum_{j\in \text{joints}}|q_j-q^{\text{default}}_j|^2$          & -0.01               \\
Collision                 & $\sum_{i\in \text{contact}}\mathbf{1}\{F_i>0.1\}$         & -10.0               \\
Stumble                   & $\mathbf{1}\{\exists i,|F^{xy}_i|>4|F^z_i|\}$      & -1.0                \\
Feet edge                 & $\sum_{i\in \text{feet}}c_i\cdot\sum_{d\in \text{dist}} \omega_d E_d[p_i]$     & -1.0                \\ 
\Xhline{0.75pt}
\end{tabular}
\end{center}
\vspace{-13pt}
\end{table}

\subsubsection{I-E Estimator} 

Previous works \cite{wang2024toward,luo2024pie} have shown that robots can infer their state and the surroundings implicitly or explicitly from proprioception and exteroception. This is critical for making full use of observations and improving adaptability. Therefore, the proposed framework employs a multi-head autoencoder-based I-E Estimator to estimate the robot's state and privileged environmental information. The estimator takes as input temporal proprioception observations $o^{H_1}_t$ ($H_1 = 10$) and local terrain reconstruction $\hat H^{\text{map}}_t$. These inputs are processed by a GRU and a CNN encoder respectively to extract proprioception features $z^{\text{prop}}_t$ and terrain features $z^{\text{map}}_t$. A set of transformer encoders is employed to fuse multimodal features and generate cross-modal estimated vectors \cite{yang2021learning}, which are then fed into the actor policy. Since each terrain feature token corresponds to a specific local zone, self-attention enables the policy to perform spatial reasoning and focus attention on relevant areas.

Among the estimated vectors, the base velocity $\hat v_t$ is explicitly estimated to help the robot be aware of its motion. Encoding $\hat z^b_t$ is decoded into the heightmap around the body $\hat H^b_t$ by an MLP decoder to enhance the policy's understanding of complex terrains. We also adopt a VAE structure to extract implicit estimation $z_t$ and predict future proprioception $o_{t+1}$ from estimated vectors. These estimations follow the configurations in \cite{luo2024pie}, and their ability to encapsulate rich privileged information into extracted features has been validated. In contrast, our framework also estimates the heightmap within 0.1m around each foot rather than only foot clearance. Encoding $\hat z^f_t$ is similarly decoded into the heightmap around the four feet $\hat H^f_t$ to capture information of footstep positions and the surrounding terrain, guiding the policy to select appropriate and edge-avoiding foot placements. 

\subsection{Terrain Reconstructor} 

The terrain reconstructor introduces local heightmap reconstruction as an explicit supervision target for extracting features from depth images. This forces the vision encoder to fully capture terrain features necessary for safe locomotion from view-limited and information-sparse egocentric vision, while also providing simple exteroception with clear features to the locomotion policy. The reconstructor receives proprioception features $z^\text{prop}_t$ and temporal depth images $d^{H_2}_t$ ($H_2=2$) as inputs, and reconstructs an accurate heightmap surrounding the robot's body in its local frame. The reconstructed heightmap spans an area from 0.5m behind the robot to 1.1m in front, with a width of 0.8m and a resolution of 5cm. The reconstruction results in the real world and simulation are shown in Fig. \ref{fig1}-B.

\begin{figure}[!tbp]
  \centering
  \vspace{3pt}
  \includegraphics[width=1.0\linewidth]{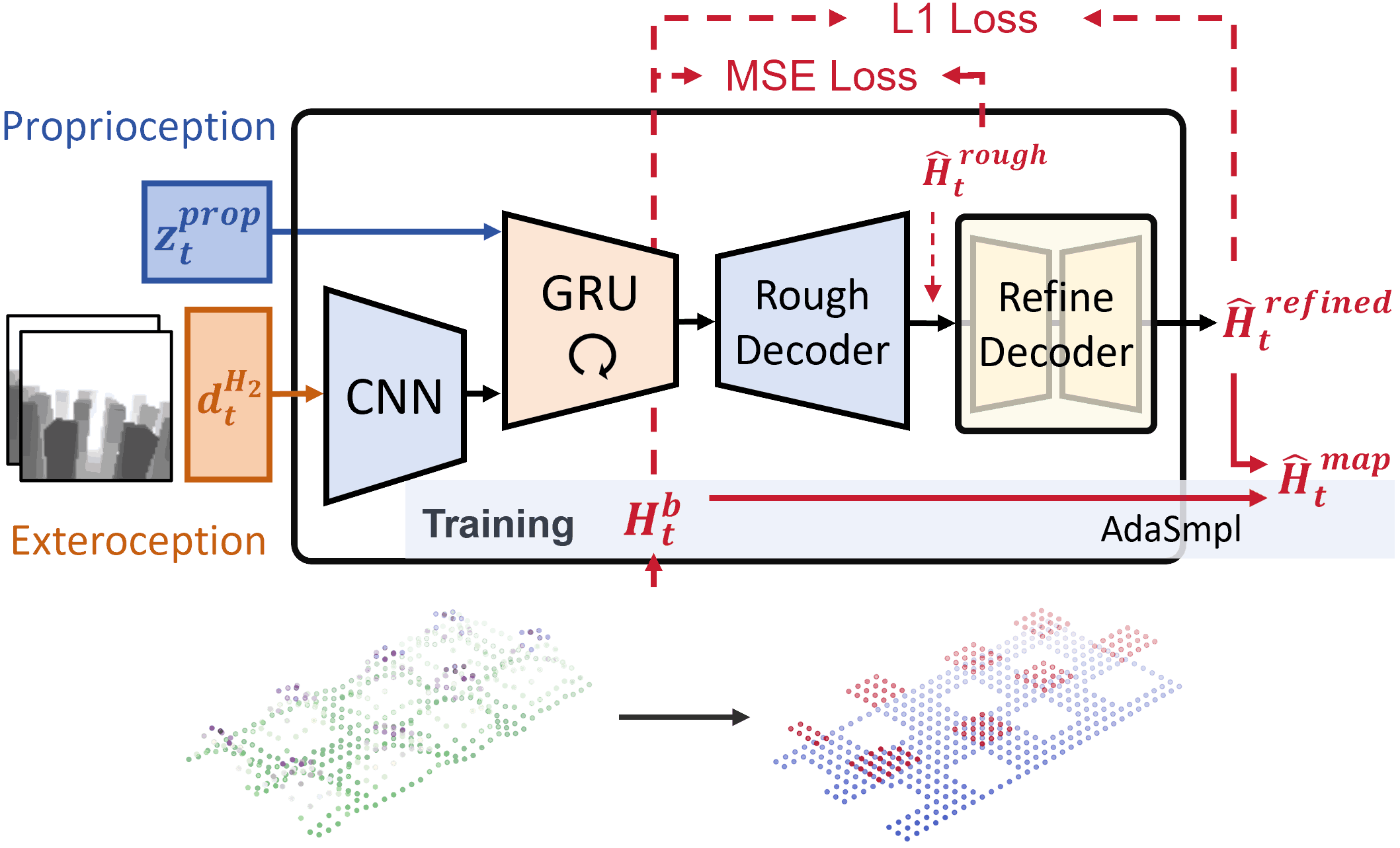}
  \caption{The terrain reconstructor reconstructs and refines the local terrain heightmap, including regions beneath and behind the robot, from proprioception features and depth images. During training, we employ AdaSmpl to reduce the learning difficulty of locomotion policy in the early stages.}
  \label{fig3}
  \vspace{-13pt}
\end{figure}

As shown in Fig. \ref{fig3}, the terrain reconstructor uses a CNN encoder to extract visual features from depth images $d^{H_2}_t$ and are concatenated with $z^{\text{prop}}_t$ that contain the robot's motion state. To memorize information about the currently invisible terrain beneath and behind the robot, the concatenated features are passed to a GRU encoder, whose output is then processed by an MLP decoder to reconstruct a rough heightmap $\hat{H}^{\text{rough}}_t$. The rough heightmap is optimized by MSE loss with respect to the ground truth $H^{b}_t$ for reconstruction loss. Recent study \cite{duan2024learning} has shown that the rough heightmap generated via only one-stage reconstruction often exhibit limited feature representation, leading to substantial noise and non-flat surfaces around the edges. To address this, we employ a U-Net-based decoder \cite{ronneberger2015u} to eliminate noise and produce a refined heightmap $\hat{H}^{\text{refined}}_t$, which is optimized using L1 loss. This refinement process improves the accuracy of edges and flat surfaces difficult to capture during rough reconstruction, thus easing feature extraction for the subsequent module. All components of the terrain reconstructor are trained together with the locomotion policy.

\subsubsection{Adaptive Sampling}

In the early stages of training, the reconstruction output often exhibits significant noise and limited terrain information, preventing locomotion policy from developing an accurate environmental understanding. To address this, we introduce an adaptive sampling (AdaSmpl) method, where the heightmap ground truth is used as the input of locomotion policy with a certain probability during training rather than reconstructed results. This probability is adaptively adjusted based on training performance:
$$
\setlength{\abovedisplayskip}{5pt}
\setlength{\belowdisplayskip}{5pt}
p_{\text{smpl}}=\text{tanh}(CV(R)) \eqno{(6)}
$$
where $CV$ is the coefficient of variation of episode reward $R$. A large $CV$ requires sampling from the ground truth with clear features to help the policy quickly capture critical terrain information, reducing the difficulty of obtaining sparse rewards in complex terrains. Conversely, a small $CV$ suggests good learning progress, with the policy gradual transitioning to reconstructed results. This also provides the policy with robustness against noise, preventing performance degradation when transferring to the real world. During deployment, the policy exclusively takes the reconstructed terrain as input.

\subsection{Training Environment}

\subsubsection{Terrain and Terrain Progressive Curriculum}

We create four types of terrain: stepping stones, balance beams, stepping beams and gaps, with the first three being sparse foothold terrains. Each type has 10 difficulty levels, and increases using the terrain curriculum from \cite{rudin2022learning}. Detailed terrain settings are presented in Appendix\ref{appendixB}.

\begin{figure}[!tbp]
  \centering
  \vspace{3.5pt}
  \includegraphics[width=0.99\linewidth]{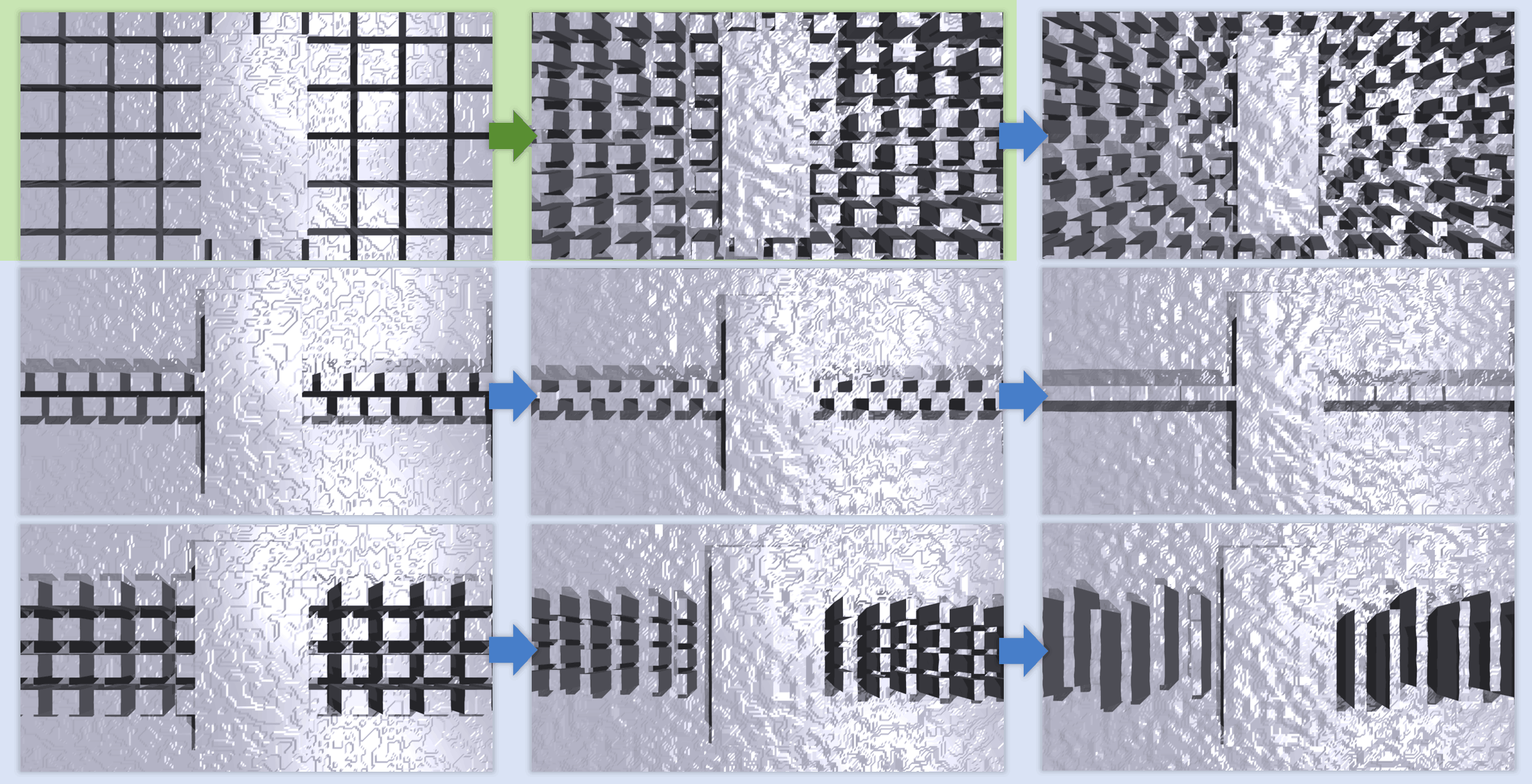}
  \caption{The robots learn basic locomotion skills on the lower-randomness stepping stones (light green) at start, and are progressively transitioned to more challenging terrains with sparse footholds (light blue).}
  \label{fig4}
  \vspace{-13pt}
\end{figure}

Due to sparse rewards and conflicting optimal gait patterns across various risky terrains with sparse footholds, it is challenging for robots to learn a highly adaptive policy from scratch. Inspired by \cite{zhang2023learning}, we developed an automatic terrain progressive curriculum to address exploration difficulties. We categorize the stepping stones into two sets: lower-randomness and higher-randomness. Furthermore, balance beams and stepping beams are designed as progressive transitions from the stepping stones, as shown in Fig. \ref{fig4}. Initially, all robots on the three sparse foothold terrains are placed on lower-randomness stepping stones and auxiliary flat terrains to learn basic locomotion skills. As training progresses, they are probabilistically assigned to balance beams, stepping beams, and stepping stones with higher sparsity and randomness. This probability follows a linear curriculum: $p_{\text{prog}}=\text{max}(\text{min}(\frac{p_{\text{max}}(T-T_{\text{start}})}{T_{\text{end}}-T_{\text{start}}}, 0), p_{\text{max}})$. Eventually, the policy learns adaptive locomotion skills across all challenging terrains.

\subsubsection{Command Sampling}

Since the terrains are aligned along the x axis, the forward velocity command is sampled within the range of $[0.0, 1.5]\text{m/s}$ in the body frame, with the lateral velocity command fixed at 0. To enable the robot's turning ability, we clamp commands below 0.3m/s to 0 and, when the robot has a command of 0, sample the angular velocity command from $[-1.2, 1.2]\text{rad/s}$. The trained policy supports both forward walking and turning.

\begin{figure*}[!htbp]
  \centering
  \vspace{3.5pt}
  \includegraphics[width=0.975\linewidth]{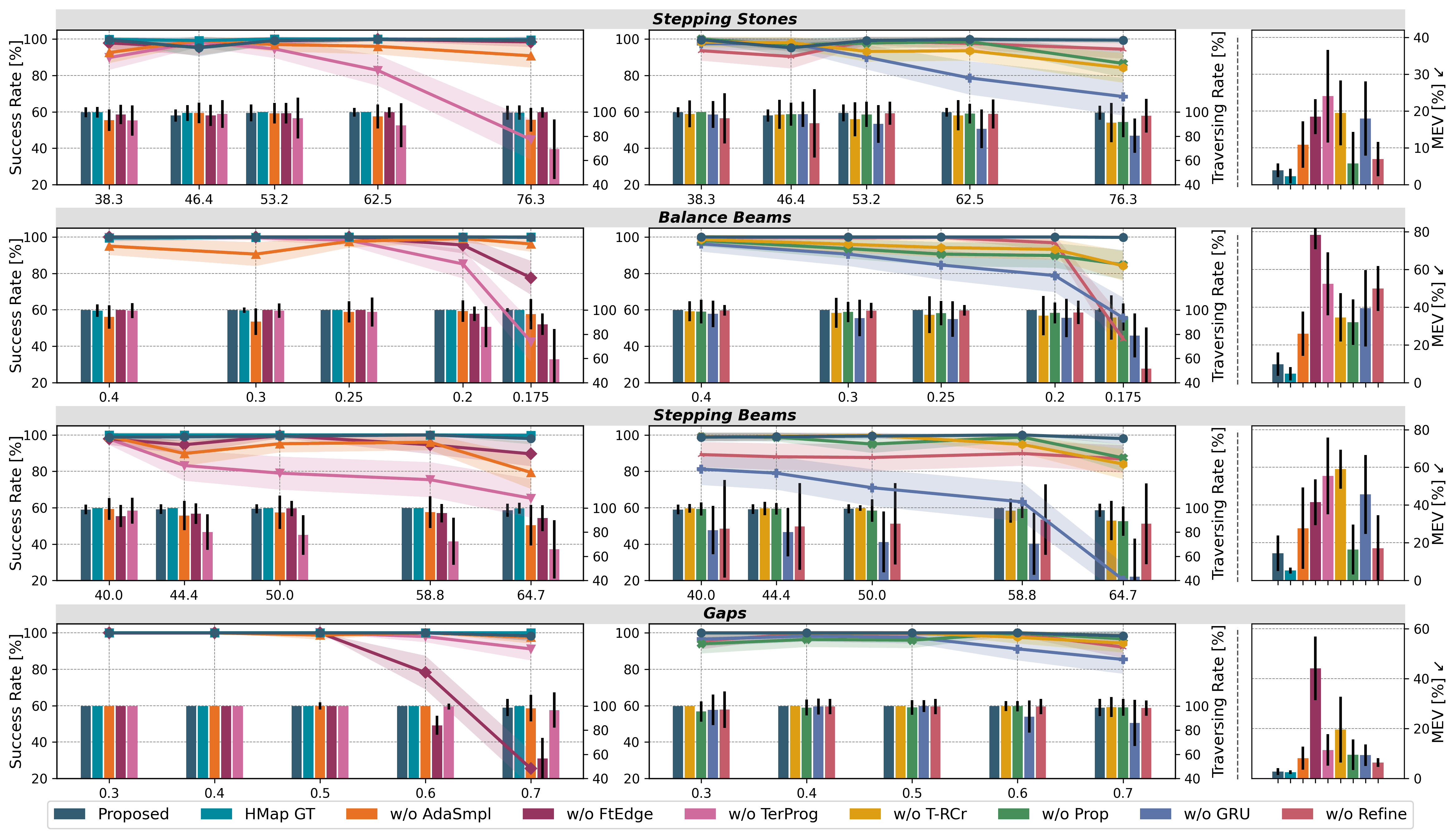}
  \vspace{-5pt}
  \caption{Ablation studies in simulation. The left and middle columns record the \textbf{success rate} (line) and \textbf{traversal rate} (bar) for training pipeline design and model architecture design. The x-axis represents terrain difficulty, defined by sparsity (with maximum randomization) for stepping stones and stepping beams, and by beam and gap width for balance beams and gaps. The right column shows the \textbf{mean edge violations (MEV)} of each policy at the highest difficulty level.}
  \label{fig5}
  \vspace{-18pt}
\end{figure*}

\section{EXPIREMENT}

\subsection{Experimental Setup}

\subsubsection{Robot and Simulation}

The proposed method was deployed on a DEEP Robotics Lite3 robot with 12 joints. The robot uses an Intel RealSense D435i camera mounted on its head to capture raw depth images at 10 Hz. The network inference is performed using TensorRT on an onboard Jetson Orin NX, with action output at 50 Hz. Training and simulation environments are implemented in IsaacGym \cite{makoviychuk2021isaac}.

\subsection{Ablation Studies}

We conducted eight ablation studies for a comparative evaluation. The first set tests the design of training pipeline:

\begin{itemize}
\item \textbf{HMap GT \cite{zhang2023learning}}: The policy takes the heightmap ground truth as input without utilizing terrain reconstructor, representing the upper bound of performance. Also serves as a comparison to works obtaining heightmap via mapping or external MoCap like \cite{zhang2023learning}.
\item \textbf{w/o AdaSmpl}: The policy trained without AdaSmpl, using reconstructed terrain as input at all times.
\item \textbf{w/o FtEdge}: The policy trained without $r_{\text{feetedge}}$.
\item \textbf{w/o TerProg}: The policy trained directly on all highly risky terrains without terrain progressive curriculum.
\end{itemize}

The second set tests the design of model architecture:

\begin{itemize}
\item \textbf{w/o T-RCr \cite{luo2024pie}}: Process depth images by a 2D-GRU instead of terrain reconstructor, retaining exteroceptive memory without heightmap reconstruction. Also serves as a comparison with previous work \cite{luo2024pie}, with slight adjustments\footnote{Involves relocating the GRU from the transformer's output to proprioception and vision encoders, with little impact on performance.} to network architecture to ensure fairness.
\item \textbf{w/o Prop}: The terrain reconstructor doesn't receive proprioception and relies solely on depth images.
\item \textbf{w/o GRU}: Replace GRU in terrain reconstructor with an MLP encoder to disable terrain feature memorization.
\item \textbf{w/o Refine}: The terrain reconstructor output the rough heightmap to the locomotion policy without refinement.
\end{itemize}

Since previous works \cite{wang2024toward,luo2024pie} have fully evaluated the importance of implicit-explicit estimation, this component is not included. Additional ablation for the estimation of heightmap around feet w/o $\hat H^f_t$ is provided in Appendix\ref{appendixC}.

\subsection{Simulation Experiments}

For each terrain, we conducted three sets of metrics for evaluating performance of the policies: 1) Success rate and traversing rate (the ratio of distance traveled before falling relative to total distance) for crossing 6m at different difficulty levels; 2) Mean edge violation (MEV, the ratio of steps landing on terrain edges to the total number of steps taken) and 3) terrain reconstruction loss while crossing 6m at the highest difficulty level. All experiments were conducted with 500 robots in 10 randomly generated terrains.

\begin{table}[!tbp]
\setlength{\abovecaptionskip}{0pt}
\setlength{\belowcaptionskip}{0pt}
\vspace{2.5pt}
\caption{Terrain reconstruction loss for various terrain reconstructors at the highest terrain difficulty level.}
\vspace{-3pt}
\label{table2}
\begin{center}
\renewcommand\arraystretch{1.22}
\begin{tabular}{lllll}
\Xhline{0.75pt}
\multicolumn{1}{c}{\multirow{2}{*}{Terrain}} & \multicolumn{4}{c}{Terrain Reconstruction Loss (MAE) {[}cm{]}} \\
\cline{2-5} 
\multicolumn{1}{c}{} & \textbf{Proposed}  & w/o Prop  & w/o GRU    & w/o Refine \\
\hline
Stepping Stones      & \textbf{5.21±0.57} & 6.56±0.64 & 9.45±1.47  & 6.83±0.61  \\
Balance Beams        & \textbf{4.09±0.34} & 4.46±1.38 & 6.31±1.33  & 5.38±1.13  \\
Stepping Beams       & \textbf{8.29±1.15} & 9.42±3.83 & 10.76±4.29 & 8.95±2.62  \\
Gaps                 & \textbf{2.39±0.75} & 3.28±2.30 & 5.09±0.61  & 2.71±0.19  \\
\Xhline{0.75pt}
\end{tabular}
\end{center}
\vspace{-22pt}
\end{table}

\subsubsection{Terrain Reconstruction Accuracy}

We evaluated the accuracy of terrain reconstruction, as shown in Table \ref{table2}. The proposed terrain reconstructor achieves superior performance in all complex terrains. However, w/o Prop, which neglects important motion states (e.g. base velocity) from proprioception features, is hard to learn the relationship between recent observations and previous reconstructions using visual inputs alone, resulting in an inability to accurately update the memory. w/o GRU cannot memorize historical terrain features, relying solely on information from the recent two frames. As a result, it struggles to reconstruct the terrain beneath and behind the robot, leading to the highest reconstruction loss. w/o Refine lacks the refining process and, although capable of roughly indicating stone and beam positions, fails to clearly represent the edges. These results demonstrate that precise terrain reconstruction requires a combination of proprioception, vision, memory, and refinement.

It should be noted that we focus on how terrain reconstruction enhances locomotion, rather than on mapping itself. By incorporating it as a supervision task within the vision encoding pipeline, we establish an effective intermediate for visual feature extraction and exteroception-aided environmental understanding. Compared to works like [1, 13, 35], we use low-cost single-view depth images as input, avoiding the complexities of mapping and global estimation.

\subsubsection{Policy Performance}

\begin{figure}[!tbp]
  \centering
  \vspace{3pt}
  \includegraphics[width=0.97\linewidth]{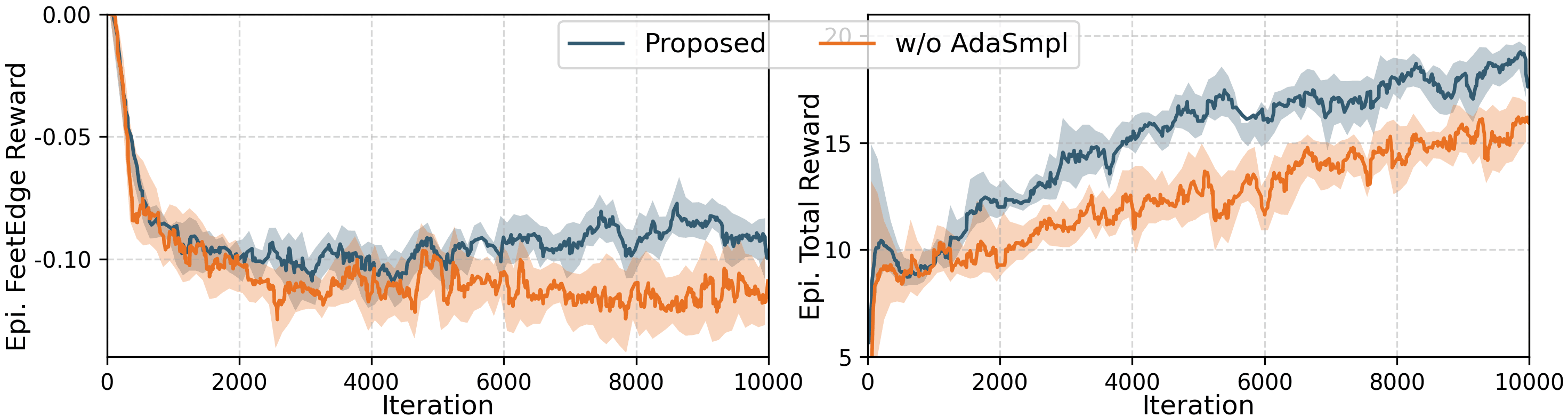}
  \vspace{-3pt}
  \caption{Plots of $r_{\text{feetedge}}$ and total reward in an episode for the proposed method and w/o AdaSmpl, where the locomotion policy receives under-trained reconstructed heightmap in the early stage instead of the ground truth with clear features. This prevents it from quickly extracting and learning to use critical $r_{\text{feetedge}}$-like rewards related terrain information, leading to a decline in both reward and performance.}
  \label{fig6}
  \vspace{-18pt}
\end{figure}

As shown in Fig. \ref{fig5}, owing to an efficient training pipeline and a well-structured model architecture, the proposed framework achieves exceptional performance across all challenging terrains. It outperforms other policies in all metrics and closely approximates the upper-bound HMap GT \cite{zhang2023learning}, which benefits from access to externally obtained privileged heightmap ground truth. In contrast, w/o AdaSmpl exhibits a decrease in performance, as its ability to extract effective terrain features (e.g. edges) is hindered by noisy and information-poor reconstruction during early training. This, in turn, increases the difficulty of obtaining sparse rewards, as shown in Fig. \ref{fig6}. The performance of w/o FtEdge and w/o TerProg declines significantly. The former often places feet on the edges, causing instability or falls, resulting in a high MEV metric. The latter struggles to learn optimal actions as it trains directly on all challenging terrain types from scratch, which make it hard to explore actions progressively, leading to complete failure on risky terrains. These results emphasize the importance of effective reward mechanisms and progressive training pipelines that facilitate gradual development of a comprehensive environmental understanding and generation of precise actions.

In terms of model architecture, w/o T-RCr \cite{luo2024pie} and w/o Prop yield similar results as both essentially use a GRU to process depth images. However, w/o T-RCr only represents visual features implicitly without incorporating terrain reconstruction, making it harder to extract sufficient terrain information necessary for safe locomotion from information-sparse images. This leads to a higher MEV metric. w/o Prop and w/o GRU lack motion state knowledge or memory mechanisms, which prevents them from reproducing terrain information accurately, both resulting in reduced effectiveness compared to the proposed policy. However, w/o GRU performs worse across all terrains due to its inability to memorize critical terrain features, leading to frequent missteps. For w/o Refine, although the refinement does not contribute additional information, the single rough reconstruction fails to represent edges clearly, leading to higher MEV metric and lower success rates on edge-dependent terrains like balance beams. Only accurate terrain reconstruction with clear features allows the policy to extract essential terrain details sufficiently from limited onboard inputs, enabling stable locomotion on sparse footholds.

\subsection{Real-World Experiments}

\begin{figure*}[!htbp]
  \centering
  \includegraphics[width=0.99\linewidth]{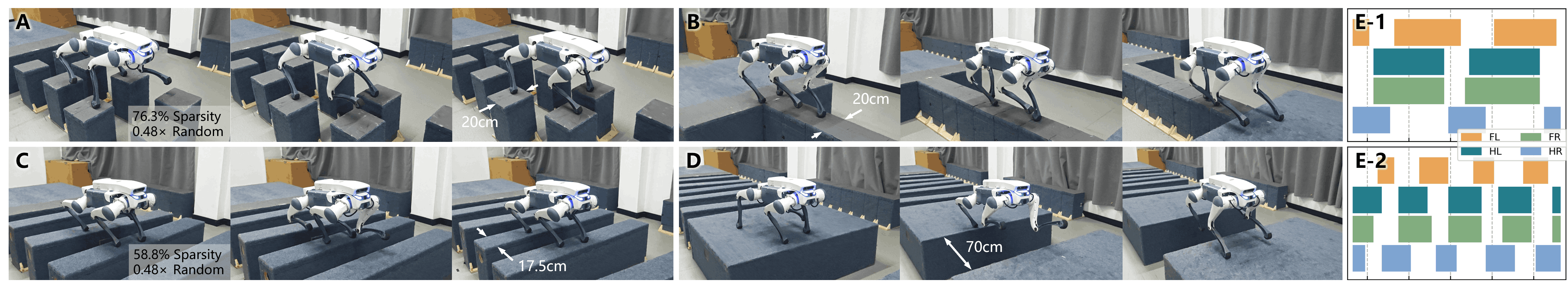}
  \vspace{-3pt}
  \caption{(A-D): Key frames of the robot traversing terrains with sparse footholds in the real world, highlighting outstanding locomotion and generalization capabilities. (E): The robot exhibits different gait patterns on stepping stones and balance beams. (A)(E-1): Traversing stones slowly with a low step frequency to ensure that each foot lands in the safe area away from the edges. (B)(E-2): First aligning the body with the center of the beam, then rapidly shifting the legs into a dynamically stable inverted triangular stance and crossing the balance beam with a high step frequency to counteract external disturbances.}
  \label{fig7}
  \vspace{-16pt}
\end{figure*}

As shown in Table \ref{table3} and Fig. \ref{fig7}, we evaluated ablation policies on real-world terrains similar to those in the simulation (excluding HMap GT due to the lack of a MoCap device). The proposed framework consistently demonstrated robust performance across all terrains, highlighting its excellent sim-to-real transferability that outperforms all other policies. w/o FtEdge exhibited a tendency to place feet near terrain edges, leading to complete failures on gaps due to unstable slips or missing the edges. w/o TerProg, which was trained directly on challenging sparse terrains, struggled to discover highly adaptive policies, resulting in unrealistic actions in the real world. Surprisingly, the performance of w/o T-RCr deteriorated significantly in real-world tests, often leading to missteps, edge placements, or directional deviations that caused falls. This suggests that terrain reconstruction is crucial for effectively capturing and accurately representing necessary terrain information during visual feature extraction, ensuring safe foot placements.

\begin{table*}[!htbp]
\setlength{\abovecaptionskip}{0pt}
\setlength{\belowcaptionskip}{0pt}
\vspace{3pt}
\caption{Ablation studies in the real world. Each policy was tested 5 times with success rate and traversing rate recorded.}
\vspace{-3pt}
\label{table3}
\begin{center}
\renewcommand\arraystretch{1.22}
\begin{tabular}{llllllllll}
\Xhline{0.75pt}
\multicolumn{1}{c}{\multirow{2}{*}{Terrain}} &
\multicolumn{1}{c}{\multirow{2}{*}{Metrics}} &
\multicolumn{8}{c}{Real World Results} \\ 
\cline{3-10} 
\multicolumn{1}{c}{} &
\multicolumn{1}{c}{} &
\textbf{Proposed} & w/o AdaSmpl & w/o FtEdge & w/o TerProg & w/o T-RCr & w/o Prop & w/o GRU & w/o Refine \\ 
\hline
\multirow{2}{*}{\shortstack{Stepping Stones\\ \emph{\footnotesize{76.3\% sparsity}}}} & Suc. Rate & \textbf{1.0}       & 0.8       & 0.6       & 0.2       & 0.2       & 0.4       & 0.2      & 0.4    \\
& Trav. Rate & \textbf{1.00±0.00} & 0.97±0.06 & 0.75±0.34 & 0.55±0.37 & 0.59±0.25 & 0.89±0.19 & 0.53±0.28 & 0.74±0.27 \\
\multirow{2}{*}{\shortstack{Balance Beams\\ \emph{\footnotesize{0.2m width}}}}   & Suc. Rate    & \textbf{0.8}       & 0.6       & 0.4       & 0.4       & 0.4       & 0.4       & 0.2      & 0.6       \\
& Trav. Rate & \textbf{0.99±0.03} & 0.89±0.16 & 0.78±0.21 & 0.87±0.14 & 0.74±0.20 & 0.75±0.19 & 0.74±0.25 & 0.92±0.15 \\
\multirow{2}{*}{\shortstack{Stepping Beams\\ \emph{\footnotesize{58.8\% sparsity}}}}  & Suc. Rate    & \textbf{1.0}       & 0.8       & 0.4       & 0.6       & 0.4       & 0.6       & 0.2      & 0.4       \\
& Trav. Rate & \textbf{1.00±0.00} & 0.99±0.03 & 0.78±0.31 & 0.80±0.28 & 0.62±0.39 & 0.85±0.25 & 0.42±0.33 & 0.71±0.34 \\
Gaps \ \emph{\footnotesize{0.7m width}}  & Suc. Rate    & \textbf{1.0}       & 0.8       & 0.0       & 0.4       & 0.2       & 0.6       & 0.4      & 0.2       \\ 
\Xhline{0.75pt}
\end{tabular}
\end{center}
\vspace{-14pt}
\end{table*}

\section{CONCLUSION}

Accurate and comprehensive environmental perception is essential for traversing risky terrains with sparse footholds. In this work, we propose a novel single-stage end-to-end RL-based framework that, for the first time, uses only low-cost onboard depth images to achieve highly agile and adaptive locomotion on highly sparse and random terrains, and demonstrates successful zero-shot transfer to real-world scenarios. By utilizing local heightmap as an intermediate reprentation, our framework enables effective extraction of terrain features crucial for safe foot placement from view-limited egocentric vision, while also alleviating the challenges of exploring and generating precise action in complex terrains. These results further extend the potential applications of quadrupedal robots across various environments.

This work validates the efficacy of the proposed method in environmental perception and locomotion on sparse footholds. Although only four typical terrains were evaluated, this approach has the potential to be extended to more generalized terrains (e.g. stairs or stepping stones with height variations). In the future, we aim to further improve the policy's applicability and motor performance by training in more varied environments or using real-world data.




\setcounter{section}{0}
\renewcommand{\thesection}{\Alph{section}}
\section*{APPENDIX}

\subsection{Computational Details of $r_{\textnormal{feetedge}}$}
\label{appendixA}

$E_d[p_i]$ in $r_{\text{feetedge}}$ is a boolean mask, with a size matching the entire terrain, that indicates whether foot position in the world frame $p_i$ is within the specified distance threshold from the edge. We select two distances, $\text{dist}=[2.5, 5.0]\text{cm}$, and corresponding penalty coefficients, $\omega_{d}=[1.0, 0.5]$. When converting the terrain from heightfield to trimesh, vertical edges are detected by checking if the slope exceeds the threshold. Then we expand the edge outward using a morphological dilation operation to create the mask $E_d$. The outer mask excludes regions that overlap with the inner one.

\subsection{Terrain Setup Details}
\label{appendixB}

To enhance the policy's adaptability, we randomly sample terrain depth from $[0.3m, 0.7m]$, and apply height variations of up to ±5 cm to all stones or beams.

\noindent\textbf{Stepping Stones.} Stone size: $[0.5\to 0.2]$m, gap (x): $[0.175\to 0.25]$m, gaps (y): $[0.1\to 0.175]$m. Randomization magnitude of a stone (x/y): $[0.0\to 0.48]\times \text{gap (x/y)}$, and an entire row along x axis: $[0.0\to 0.48]\times \text{gap (x)}$. 


\noindent\textbf{Balance Beams.} In training, the balance beams start from two rows of offset stones and progressively form a single row by reducing the spacing and width of stones, as shown in Fig. \ref{fig4}. The beam width at the highest difficulty is 0.175m.


\noindent\textbf{Stepping Beams.} In training, the stepping beams start from four rows of stones and progressively form a sequence of beams by decreasing the width of stone along x axis and the spacing along y axis, as shown in Fig. \ref{fig4}. The beam width and gap at the highest difficulty is 0.15m and 0.275m.


\noindent\textbf{Gaps.} Gap width: $[0.1\to 0.7]$m.

For balance beams and stepping beams, the beam width and gap are adjusted directly during experiment.

\subsection{Additional Ablation Study on $\hat H^f_t$ Estimation}
\label{appendixC}

To further compare with previous work \cite{luo2024pie}, we conducted an additional ablation w/o $\hat H^f_t$, replacing the estimation of heightmap around feet with foot clearance. As shown in Table \ref{table4}, w/o $\hat H^f_t$ exhibits higher MEV metric. This suggests that estimating the heightmap around each foot is essential for the policy to understand the terrain around the foothold, thereby enabling it to actively avoid the terrain edges.

\begin{table}[!hbp]
\setlength{\abovecaptionskip}{0pt}
\setlength{\belowcaptionskip}{0pt}
\vspace{-2pt}
\caption{Ablation for w/o $\hat H^f_t$. Each policy was tested 500 times at the highest terrain difficulty level in simulation.}
\vspace{-3pt}
\label{table4}
\begin{center}
\renewcommand\arraystretch{1.06}
\begin{tabular}{lllll}
\Xhline{0.75pt}
\multicolumn{1}{c}{\multirow{2}{*}{Terrain}} & \multicolumn{2}{c}{$\text{Success Rate} \nearrow $} & \multicolumn{2}{c}{$\text{MEV} \searrow $} \\
\vspace{-13pt} \\
\cmidrule(r){2-3} \cmidrule(r){4-5} 
\vspace{-13pt} \\
\multicolumn{1}{c}{} & \textbf{Proposed}  & w/o $\hat H^f_t$  & \textbf{Proposed} & w/o $\hat H^f_t$ \\
\hline
Stepping Stones      & \textbf{0.99±0.00} & 0.90±0.01 & \textbf{0.04±0.02}  & 0.19±0.08  \\
Balance Beams        & \textbf{1.00±0.00} & 0.95±0.01 & \textbf{0.10±0.06}  & 0.18±0.10  \\
Stepping Beams       & \textbf{0.98±0.01} & 0.92±0.01 & \textbf{0.14±0.10}  & 0.52±0.19  \\
Gaps                 & \textbf{0.98±0.01} & 0.98±0.01 & \textbf{0.03±0.01}  & 0.21±0.04  \\
\Xhline{0.75pt}
\end{tabular}
\end{center}
\vspace{-20pt}
\end{table}

\subsection{Policy Architectures and Input Composition Details}
\label{appendixD}

\begin{table}[!htbp]
\setlength{\abovecaptionskip}{0pt}
\setlength{\belowcaptionskip}{0pt}
\vspace{-13pt}
\caption{Detailed I/O and parameters of all components.}
\vspace{-6pt}
\label{table5}
\begin{center}
\renewcommand\arraystretch{1.27}
\begin{tabular}{lll}
\Xhline{0.75pt}
\multicolumn{1}{c}{\multirow{7}{*}[-0.2ex]{T-RCr}}
& \multicolumn{1}{c}{\multirow{3}{*}[-0.3ex]{\shortstack{Depth\\Enc.}}}
& \multicolumn{1}{c}{\cellcolor{gray!15}${z^{\text{depth}}_t=GRU([MLP(z^{\text{prop}}_t),CNN(d^{H_2}_t)])}$}         \\
\hhline{>{\arrayrulecolor{black}}~~>{\arrayrulecolor{black}}-}
& & \multicolumn{1}{c}{\emph{CNN}: \{[32,64,64],[7,7,3,3],[4,4,2,2],[3,3,1,1]\}}         \\ [-3pt]
& & \multicolumn{1}{c}{\emph{MLP}: \{[64,64], ELU\}, \emph{GRU}: \{1,128\}}                          \\
\hhline{>{\arrayrulecolor{black}}~>{\arrayrulecolor{black}}--}
& \multicolumn{1}{c}{\multirow{2}{*}[-0.2ex]{\shortstack{Rough\\Dec.}}}
& \multicolumn{1}{c}{\cellcolor{gray!15}${\hat H^{\text{rough}}_t=MLP(z^{\text{depth}}_t)}$}         \\
\hhline{>{\arrayrulecolor{black}}~~>{\arrayrulecolor{black}}-}
& & \multicolumn{1}{c}{\emph{MLP}: \{[128,256], ELU\}}         \\
\hhline{>{\arrayrulecolor{black}}~>{\arrayrulecolor{black}}--}
& \multicolumn{1}{c}{\multirow{2}{*}[-0.2ex]{\shortstack{Refine\\Dec.}}}
& \multicolumn{1}{c}{\cellcolor{gray!15}${\hat H^{\text{refined}}_t=UNet(\hat H^{\text{rough}}_t)}$}         \\
\hhline{>{\arrayrulecolor{black}}~~>{\arrayrulecolor{black}}-}
& & \multicolumn{1}{c}{Shown in Fig. \ref{fig8}.}         \\
\hhline{>{\arrayrulecolor{black}}>{\arrayrulecolor{black}}---}

\multicolumn{1}{c}{\multirow{10}{*}[-0.2ex]{\shortstack{I-E\\Est.}}}
& \multicolumn{1}{c}{\multirow{2}{*}[-0.2ex]{\shortstack{Prop.\\Enc.}}}
& \multicolumn{1}{c}{\cellcolor{gray!15}${z^{\text{prop}}_t=GRU(MLP(o^{H_1}_t))}$}         \\
\hhline{>{\arrayrulecolor{black}}~~>{\arrayrulecolor{black}}-}
& & \multicolumn{1}{c}{\emph{MLP}: \{[256,128], ELU\}, \emph{GRU}: \{1,128\}}         \\
\hhline{>{\arrayrulecolor{black}}~>{\arrayrulecolor{black}}--}
& \multicolumn{1}{c}{\multirow{2}{*}[-0.2ex]{\shortstack{Hmap.\\Enc.}}}
& \multicolumn{1}{c}{\cellcolor{gray!15}${z^{\text{map}}_t=CNN(\hat H^{\text{map}}_t)}$}         \\
\hhline{>{\arrayrulecolor{black}}~~>{\arrayrulecolor{black}}-}
& & \multicolumn{1}{c}{\emph{CNN}: \{[32,64],[7,3,3],[2,1,2],[3,1,1]\}}         \\
\hhline{>{\arrayrulecolor{black}}~~>{\arrayrulecolor{black}}-}

& \multicolumn{1}{c}{\multirow{2}{*}[-0.2ex]{\shortstack{Tf.\\Enc. \cite{yang2021learning}}}}
& \multicolumn{1}{c}{\cellcolor{gray!15}${z^{\text{tf}}_t = L(SA(z^{\text{prop}}_t,z^{\text{map}}_t))}$} \\
\hhline{>{\arrayrulecolor{black}}~~>{\arrayrulecolor{black}}-}
& & \multicolumn{1}{c}{\emph{SA}: \{[256,256],64,1\}, \emph{L}: Linear}         \\
\hhline{>{\arrayrulecolor{black}}~>{\arrayrulecolor{black}}--}
& \multicolumn{1}{c}{\multirow{4}{*}{Est.}}
& \multicolumn{1}{c}{\cellcolor{gray!15}${z_t = Reparam(L(z^{\text{tf}}_t))}$, ${\hat v_t = L(z^{\text{tf}}_t)}$, ${\hat z^b_t = L(z^{\text{tf}}_t)}$}         \\ [-1pt]
& & \multicolumn{1}{c}{\cellcolor{gray!15}${\hat H^b_t = MLP(z^b_t)}$, ${\hat z^f_t = L(z^{\text{tf}}_t)}$, ${\hat H^f_t = MLP(z^f_t)}$}         \\ [-1pt]
& & \multicolumn{1}{c}{\cellcolor{gray!15}${\hat o_{t+1} = MLP([z_t,\hat v_t,\hat z^f_t,\hat z^b_t])}$}         \\
\hhline{>{\arrayrulecolor{black}}~~>{\arrayrulecolor{black}}-}
& & \multicolumn{1}{c}{\emph{MLP}:\{[64,128], ELU\}, \emph{L}: Linear}         \\ 
\hhline{>{\arrayrulecolor{black}}>{\arrayrulecolor{black}}---}

\multicolumn{1}{c}{\multirow{2}{*}[-0.4ex]{\shortstack{Asym.\\A-C}}}
& \multicolumn{1}{c}{\multirow{2}{*}[-0.2ex]{\shortstack{Actor\\Critic}}}
& \multicolumn{1}{c}{\cellcolor{gray!15}${a_t = MLP([o_t,z_t,\hat v_t,\hat z^f_t,\hat z^b_t])}$, ${V_t = MLP([o_t,s_t])}$}         \\
\hhline{>{\arrayrulecolor{black}}~~>{\arrayrulecolor{black}}-}
& & \multicolumn{1}{c}{\emph{MLP}: \{[512,256,128], ELU\}}         \\
\Xhline{0.75pt}

\multicolumn{3}{c}{\scriptsize \emph{CNN}: \{channel, kernel, stride, padding\}, \emph{MLP}: \{hidden, activation\}} \\ [-4pt]
\multicolumn{3}{c}{\scriptsize \emph{GRU}: \{n\_layer, hidden\}, \emph{SA (self-attention)}: \{hidden, token, n\_head\}}\\

\Xhline{0.75pt}

\end{tabular}
\end{center}
\vspace{-26pt}
\end{table}

\begin{figure}[!htbp]
  \centering
  \includegraphics[width=0.96\linewidth]{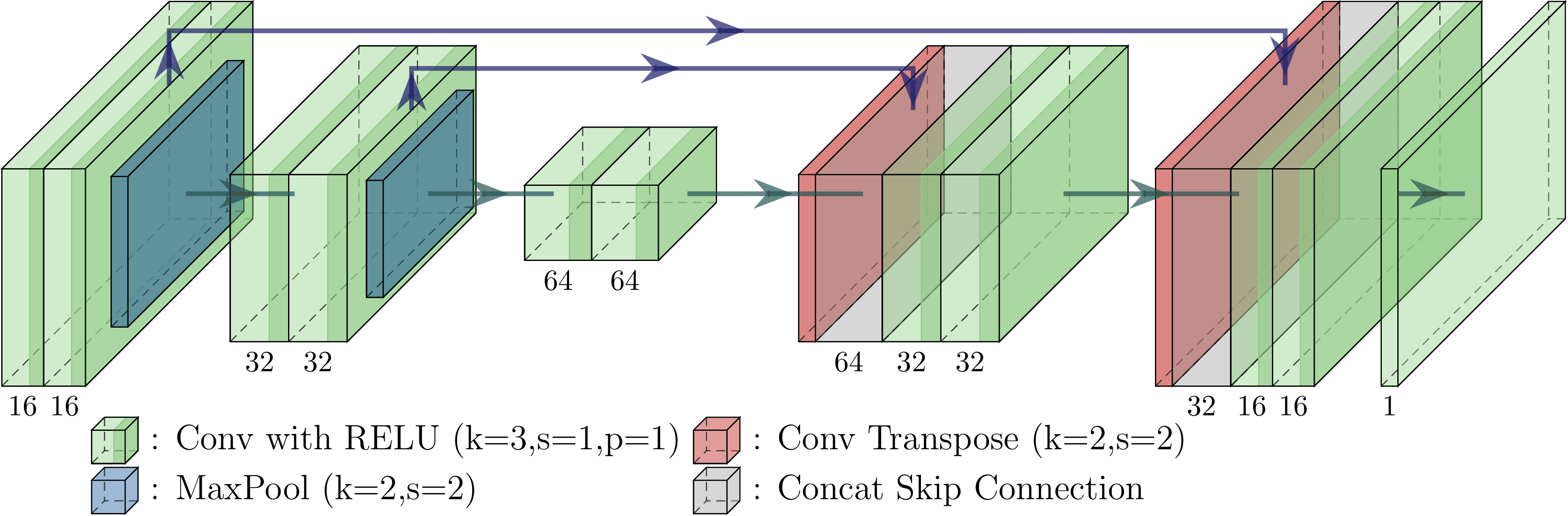}
  \vspace{-6pt}
  \caption{Architecture of the U-Net-based refine decoder.}
  \label{fig8}
  \vspace{-13pt}
\end{figure}




\bibliographystyle{IEEEtran}
\bibliography{IEEEabrv,ref}

\end{document}